\documentclass[pmlr]{jmlr}

\usepackage{longtable}

\usepackage{booktabs}
\usepackage[load-configurations=version-1]{siunitx} 

\usepackage{amsfonts}
\usepackage{enumitem}
\usepackage{breakcites} 
\usepackage{array} 
\usepackage{multicol} 

\usepackage{caption}
\usepackage{subcaption}

\theorembodyfont{\upshape}
\theoremheaderfont{\scshape}
\theorempostheader{:}
\theoremsep{\newline}

\jmlrvolume{85}
\jmlryear{2018}
\jmlrworkshop{Machine Learning for Healthcare}

\title[Sequence Transformer Networks]{Learning to Exploit Invariances in Clinical Time-Series Data using Sequence Transformer Networks}

\author{\Name{Jeeheh Oh} \Email{jeeheh@umich.edu} 
       \addr {\footnotesize Department of Electrical Engineering and Computer Science\\
       University of Michigan\\
       Ann Arbor, MI} 
       \AND
       \Name{Jiaxuan Wang} \Email{jiaxuan@umich.edu} 
       \addr {\footnotesize Department of Electrical Engineering and Computer Science\\
       University of Michigan\\
       Ann Arbor, MI}
       \AND
       \Name{Jenna Wiens} \Email{wiensj@umich.edu} 
       \addr {\footnotesize Department of Electrical Engineering and Computer Science\\
       University of Michigan\\
       Ann Arbor, MI}} 

\begin{document}

\maketitle

\begin{abstract}
Recently, researchers have started applying convolutional neural networks (CNNs) with one-dimensional convolutions to clinical tasks involving time-series data. This is due, in part, to their computational efficiency, relative to recurrent neural networks and their ability to efficiently exploit certain temporal invariances, (\textit{e.g.}, phase invariance). However, it is well-established that clinical data may exhibit many other types of invariances (\textit{e.g.}, scaling). While preprocessing techniques, (\textit{e.g.,} dynamic time warping) may successfully transform  and align inputs, their use often requires one to identify the types of invariances in advance. In contrast,  we propose the use of Sequence Transformer Networks, an end-to-end trainable architecture that learns to identify and account for invariances in clinical time-series data. Applied to the task of predicting in-hospital mortality, our proposed approach achieves an improvement in the area under the receiver operating characteristic curve (AUROC) relative to a baseline CNN (AUROC=0.851 vs. AUROC=0.838). Our results suggest that a variety of valuable invariances can be learned directly from the data.
\end{abstract}

\section{Introduction}

Clinical time-series data consist of a wide variety of repeated measurements/observations, from vitals (\textit{e.g.}, heart rate) and laboratory results to locations within a hospital \citep{che2018recurrent,lipton2015learning,oh2018generalizable}. These data vary not only in the information they encode, but also in sampling rate and number of measurements. Analogous to how certain tasks in computer vision exhibit spatial invariances, invariances frequently arise in clinical tasks involving time-series data. These invariances describe a set of transformations that, when applied to the data, magnify task-relevant similarities between examples. For example, phase invariance relates to a transformation that shifts a signal, resulting in an alignment in phase. Such transformations can be particularly useful when processing periodic signals \textit{e.g.,} electrocardiogram waveforms \citep{wiens2010active}. 

Preprocessing techniques like dynamic time warping are commonly used to exploit warping invariances and align time-series data, facilitating relevant comparisons \citep{liu2014ecg, ortiz2016heart}. However, their computational complexity (\textit{e.g.}, DTW involves solving an optimization problem for each new example) may be a factor leading to their limited use within more general settings. In addition, such approaches require \textit{a priori} knowledge of the types of invariances that are present in one's data. Due to the varied nature of clinical time-series data and their associated prediction tasks, we expect that many such tasks involve multiple invariances that may not be known beforehand. This and the fact that these invariances are likely task specific, are some of the main roadblocks in efficiently exploiting these invariances. 

To addresses these challenges, we propose Sequence Transformer Networks, an approach for learning task-specific invariances related to amplitude, offset, and scale invariances directly from the data. Our approach consists of an end-to-end trainable framework designed to capture temporal and magnitude invariances. Applied to clinical time-series data, Sequence Transformer Networks learn input- and task-dependent transformations. In contrast to data augmentation approaches, our proposed approach makes limited assumptions about the presence of invariances in the data. Learned transformations can be efficiently applied to new input data, leading to an improvement in overall predictive performance. We demonstrate the utility of the proposed approach in the context of predicting in-hospital mortality given 48 hours of data collected in the intensive care unit (ICU). Relative to a baseline that does not incorporate any transformations, Sequence Transformer Networks result in significant improvements in predictive performance. 

\paragraph{Technical Significance}
Our technical contributions are as follows: 1) we propose the use of Sequence Transformer Networks, an end-to-end trainable framework designed to capture temporal and magnitude invariances, 2) on a real data task, we evaluate the relative contribution of each individual component of Sequence Transformer Networks towards the overall performance of the network and 3) present visualizations of the types of learned invariances and investigate the effects of Sequence Transformer Networks on intra-class signal similarity. This work represents a step toward understanding and learning to exploit invariances in clinical-time series data. 

\paragraph{Clinical Relevance}
To investigate the capability of the proposed approach, we consider the task of predicting in-hospital mortality given clinical time-series data from the first 48 hours of an ICU admission. We chose to focus on this task since it is widely investigated in the machine learning for healthcare literature, facilitating comparisons with state-of-the-art. Despite its widespread use as a benchmark task \citep{harutyunyan2017multitask} and potential clinical use as an estimate of severity of illness, we recognize that a model for predicting in-hospital mortality may be of limited clinical utility. Though we consider the improvements our proposed approach offers in the context of this benchmark task, we hypothesize that our approach applies more broadly to other tasks involving clinical time-series data. 

\section{Background \& Related Work}

Tasks involving time-series data may exhibit a number of different invariances. We refer the reader to the following paper for an in-depth discussion of types of invariances present in time-series data \citep{batista2011complexity}, but for completeness include a summary of common invariances in Table \ref{tab:invariances}. To exploit these invariances, researchers often turn to neural networks. In particular, one-dimensional (1D) convolutional neural networks (CNNs), by design, efficiently exploit phase invariance. This property, in addition to their computational efficiency achieved by weight sharing, has led to their successful application to a variety of tasks involving sequential data \citep{cui2016multi, wang2015imaging,gehring2017convolutional, dauphin2016language,yin2017comparative}, and more specifically clinical time-series data \citep{razavian2015temporal, pmlr-v56-Razavian16, suresh2017clinical,bashivan2015learning}. Recognizing that clinical time-series data exhibit other types of invariance, beyond phase invariance, we propose augmenting CNNs to explicitly account for task-irrelevant variation.

In other domains, to exploit invariances researchers either i) augment their training data by applying a variety of transformations or ii) modify the neural network architecture. The first approach is most popular in domains where it is straightforward to generate realistic training examples (\textit{e.g.}, natural images). Common image invariances include rotation, scale, translation and warping. Such transformations are easily applied to existing images to create additional, realistic training examples. While less common in the healthcare domain, there have been successful examples of data augmentation for health data. For example, \cite{um2017data} augmented  multivariate time-series data collected from a wearable sensor placed on a person's wrist in order to improve monitoring of patients with Parkinson's disease. The authors applied transformations such as noise and rotations, selected based on the task. However, in general it is not straightforward to apply such data augmentation schemes to clinical data because of the large number of potential invariances. Moreover, clinical time-series data extracted from electronic health records often consist of high-dimensional data measuring many different aspects of a patient's health. This increases the complexity of identifying reasonable transformations and makes a brute-force search over possible transformations computationally intractable. 

Our work is more in-line with the second approach that does not rely on additional data. Instead, the architectures are modified to exploit a particular invariance \citep{wang2012towards, razavian2015temporal, pmlr-v56-Razavian16, forestier2017generating, wang2015imaging, cui2016multi}. For example, in \citep{razavian2015temporal} and \citep{pmlr-v56-Razavian16}, the authors tackle warping by using multiple filter sizes. More specifically, three different sized filters were used to capture a range of long- and short-term temporal patterns. These different resolutions corresponded to separate convolutional layers, combined at the final fully connected layer. \cite{cui2016multi} propose an additional preprocessing step, in which they resample and smooth their input in order to capture multiscale patterns and remove noise. Transformed versions of the inputs were treated as additional channels to the original image. Similar to \citep{razavian2015temporal,pmlr-v56-Razavian16}, this method incorporates a local convolution stage that looks at each type of transformation (none, smoothing, down-sampling) independently before combining. Both of these works are geared toward specific invariances, in this case scale invariance, and require the user to determine the different filter sizes or sampling rates. 

Recognizing the difficulty in identifying potential invariances or transformation \textit{a priori}, we focus on learning the invariances directly from the data. Our proposed approach extends work by \cite{jaderberg2015spatial}, in which a spatial transformer network is used to learn spatial invariances directly from the data. In \citep{jaderberg2015spatial}, the parameters of a spatial transformer network are learned jointly with the parameters of a CNN. The transformer network applies a learned set of transformations including affine transformations tailored to each input before passing it through a CNN. Since we focus on clinical time-series, and not images, we adapt the set of possible transformations. Specifically, our proposed method tackles amplitude and offset invariances (which we will refer to as magnitude invariance), phase invariance, and uniform scale invariance, and learns input-specific transformation parameters directly from the data. We describe the details of our approach in the next section.

\begin{table}[h]
  \centering 
\caption{A list of possible invariances summarized from \cite{batista2011complexity}. Any number or combination of invariances may arise in clinical time-series data, or time-series data in general.}
  \label{tab:invariances} 
\scalebox{0.9}{
  \begin{tabular}{  |m{3cm}|m{9cm}|}
  \toprule
    \textbf{Invariance} & \textbf{Description}  \\
    \toprule
    Amplitude & A transformation of the amplitude of the time series. This can occur when the scale or unit of measurement of two time series differs (\textit{e.g.}, temperature in Celsius vs. Fahrenheit). \\ \hline
   Offset & A transformation that uniformly increases/decreases the value of a time series. For example, two patients may have different resting heart rates.\\\hline
   \multicolumn{1}{|l|}{Local Scaling (“Warping”)} & A transformation that locally stretches or warps the duration of the time series. Local warping is often referenced in conjunction with Dynamic Time Warping (DTW), a good, established measure of similarity between time series with local scaling invariance.\\ \hline
  Uniform Scaling & A transformation that globally stretches the duration of the time series. For example, when resting heart rates differ between patients, the progression of the same temporal pattern may be consistently slower in one patient versus another.  \\\hline
    Phase & A transformation that shifts the start time of a time series. This occurs in periodic signals such as heartbeat and blood pressure waveforms. \\\hline
   Occlusion & A transformation that randomly removes data. This can arise when measurements are irregularly sampled or missing. \\ \hline
    Noise & A transformation that adds or removes noise. For example, many single point sensors are susceptible to noise that might not be indicative of the whole body's condition but indicative of that sensor's particular location.\\
    \bottomrule
  \end{tabular}
  }

\end{table}

\subsection{Problem Setup \& Notation}

We consider the application of 1D CNNs to clinical time-series data for predicting a specific outcome. Formally, given a set of $n$ labeled examples consisting of $d$ features measured at $T$ time steps ($X\in\mathbb{R}^{n\times d\times T}$) and the outcome labels $\boldsymbol{y}\in\{0,1\}^{n}$, our goal is to learn a mapping from $\{\boldsymbol{x_t^{(i)}}\}_{t=1}^{T}$ to $y^{(i)}$, where $\boldsymbol{x_t^{(i)}} \in \mathbb{R}^d$ and $i \in \{1 \cdots n\}$ is an index into the $i^{th}$ sample. The $d$ features may consist of both time-varying and time-invariant data. We represent each feature as a set of $T$ measurements. For time-varying data for which we do not have a measurement at time $t$, we carry forward the most recent value. For time-invariant data, we copy the measurement across all $T$ time-steps as in \citep{fiterau2017shortfuse}. Additional details pertaining to the specific dataset used through our experiments can be found in Section 4.

\section{Sequence Transformer}

Applied to time-series data, 1D convolutions inherently capture some invariance in the data. In particular, CNNs are capable of efficiently handling phase invariance (\textit{i.e.}, the use of a filter slid along the temporal dimension allows for variability in the starting point of temporal patterns.) CNNs also handle noise invariance, to a degree. Max pooling coupled with multiple layers allows the model to smooth the inputs and learn higher-level abstractions.  

\begin{figure}
    \centering
    \includegraphics[width=.6\linewidth]{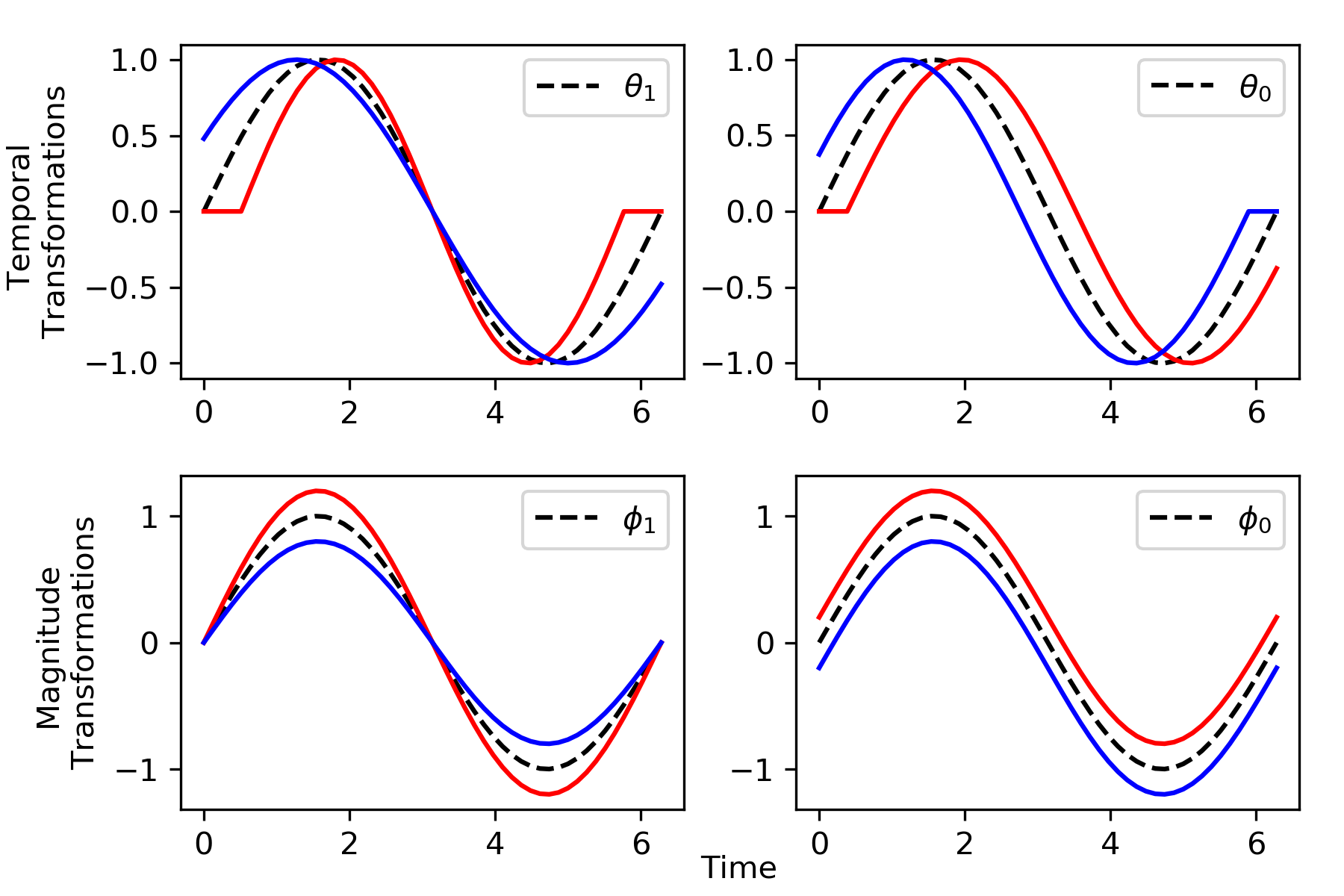}
    \caption{Examples of the types of transformations/invariances that can be learned by a Sequence Transformer Network applied to a sine wave. $\theta_1$: scaling invariance, $\theta_0$: phase invariance, $\phi_1$: amplitude invariance, $\phi_0$: offset invariance. The dashed line represents the original signal and the blue lines represent potential transformations of the signal. While CNNs can efficiently exploit phase invariance, Sequence Transformers can augment other types of architectures facilitating the capture of other types of invariances. }
    \label{STN13ak}
\end{figure}

However, there are other types of invariances that we would like to consider, in particular temporal invariance such as scaling, in addition to magnitude invariance related to the amplitude and offset of the signal. Figure \ref{STN13ak} shows examples of these types of invariances on a sine wave. Due to the inherent differences between these types of invariances, we address them separately in the two subsections that follow. For simplicity, in this section, methods are presented in terms of a univariate signal, but later our experiments focus on a multivariate application. 

\subsection{Temporal Transformations}
To capture invariance related to warping and scaling, we begin by learning to transform data along the temporal dimension. As in \citep{jaderberg2015spatial}, this stage consists of two separate pieces i) learning the transformation parameters and ii) mapping those transformations in terms of discrete data points. We discuss each, in turn, below. \\

\noindent \textbf{Transformation Network.} We begin by learning a transformation that takes points from the original input (\textit{i.e.}, the source) and maps them to a new temporal location in the target. Since we only consider linear transformations along the temporal axis, we respect the ordering of values, but can stretch, compress, flip and/or shift the signal (across the temporal axis).

\begin{align}\label{eqn:temporal_transform}
    t
    =\boldsymbol{\theta^{(i)}}
    \begin{pmatrix}
        t' \\ 1
    \end{pmatrix}
    = 
    \begin{pmatrix}
        \theta^{(i)}_1&\theta^{(i)}_0
    \end{pmatrix}
    \begin{pmatrix}
        t' \\ 1
    \end{pmatrix}
\end{align} \label{eqn_time}

Equation (\ref{eqn_time}) gives a mapping between the transformed time point $t'$ and original time point $t$. Given a univariate time-series $\{x_t^{(i)}\}_{t=1}^T$, $t$ represents the $t^{th}$ position along the temporal axis of the time-series. We learn a linear temporal transformation $\boldsymbol{\theta^{(i)}}\in\mathbb{R}^{n\times2}$ that applies to these indices. Specifically, we generate $t'$ for $t'=1,...,T'$. $T'$ represents the length of the transformed sequence and can be set to any positive integer. Here, for convenience, we set $T=T'$. The transformation parameters $\boldsymbol{\theta^{(i)}}$ are learned via a two-layer CNN that is fed inputs $\{x_t^{(i)}\}_{t=1}^{T}$. Network architecture details are outlined in Figure \ref{fig1}. Given a particular position, $t'$, in the target time series, we compute the corresponding position in the original time series and set $x_{t'}$ to refer to $x_{t=\theta_{1}t' + \theta_{0}}$.\\

\noindent \textbf{Discrete Mapping.} Since $\theta_{1}t' + \theta_{0}$ for $t'=1,...,T'$ is not guaranteed to map to a positive integer (\textit{i.e.}, an index), we require an additional step to apply the learned transformation. We complete the mapping using linear sampling, in which we take an average over the two nearest neighbors (one from left, one from the right)\footnote{Signals are padded by the last known value so there is no edge case where a point has only one neighbor.} weighted by the distance from the original transformed point.

\subsection{Magnitude Transformations}

In order to adapt to amplitude and offset invariance, we propose an additional learned transformation, one that is applied to the values instead of the coordinates. Given the temporally transformed inputs ${\{x^{(i)}_{t^{'}}\}}_{t^{'}=1}^{T^{'}}$, we apply the following linear transformation:

\begin{align}
    x_{t^{'}}^{'(i)}=
    \boldsymbol{\phi^{(i)}}\cdot x^{(i)}_{t^{'}}=
    \begin{pmatrix}
        \phi^{(i)}_1 &\phi^{(i)}_0
    \end{pmatrix}
    \cdot 
    \begin{pmatrix}
        x_{t^{'}}^{(i)}\\1
    \end{pmatrix}
\end{align}

This allows us to shift, flip, stretch, and compress the signal along its magnitude. Since this transformation applies directly to the values of the signal, we do not require a discrete mapping component. It should be noted that the transformation, $\boldsymbol{\phi^{(i)}}\in\mathbb{R}^{n\times2}$ is a function of $x$, thus it can vary from example to example.

\subsection{Sequence Transformer}
\begin{figure}
    \centering
    \includegraphics[width=1\linewidth]{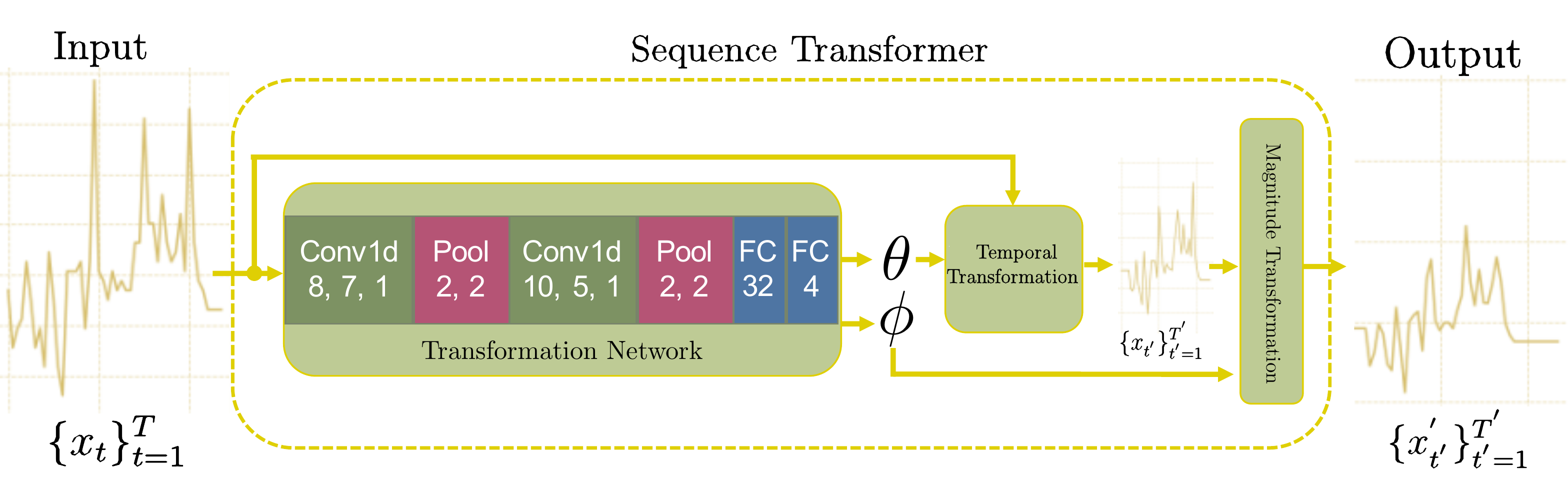}
    \caption{The architecture of a Sequence Transformer. Inputs $\{x_t\}_{t=1}^{T}$ (shown as univariate for illustration purpose) are fed into a Transformation Network that outputs transformation parameters $\boldsymbol{\theta}$ and $\boldsymbol{\phi}$. Convolutional and maxpool layers are annotated with the number of outputted channels (omitted for maxpool), filter size and stride. Fully connected layers (FC) are annotated with the number of neurons. The temporal transformation is applied via discrete mapping and the magnitude transformation is applied via linear transformation. The output  $\{x^{'}_{t'}\}_{t^{'}=1}^{T^{'}}$ represents the transformed inputs.}
    \label{fig1}
\end{figure}

We refer to the temporal transformation combined with the magnitude transformation as a Sequence Transformer (Figure \ref{fig1}). The Sequence Transformer computes both the $\boldsymbol{\theta}$ and $\boldsymbol{\phi}$ transformation parameters based on the input and applies them to the signal.

While we presented this approach in the context of a univariate signal, the technique generalizes to multivariate signals. In a multivariate setting, the Transformation Network outlined in Figure \ref{fig1} takes as input $\{\boldsymbol{x_t}\}_{t=1}^{T}$, where $\boldsymbol{x_t}\in\mathbb{R}^{d}$. The Transformation Network then estimates $[\boldsymbol{\theta},\boldsymbol{\phi}]$, based these data and the underlying model parameters. Although the model parameters are consistent across all examples, the resulting \textit{transformation} parameters (\textit{i.e.}, $\boldsymbol{\theta}\in\mathbb{R}^{n\times2}$ and $\boldsymbol{\phi}\in\mathbb{R}^{n\times2}$) are specific to each example. This transformation is then applied to all signals in the input (note that temporal transformations have no effect on time-invariant data, but these signals can still be transformed in a meaningful way). 

\section{Experimental Setup}
In this section, we describe our dataset and prediction task, the baseline CNN architecture and implementation details.

\subsection{Dataset \& Prediction Task}

To measure the utility of the proposed approach on a real dataset, we consider a standard sequence-level classification task: predicting in-hospital mortality based on the first 48 hours of data collected during an intensive care unit visit. We use data from MIMIC III \citep{johnson2016mimic}. As in \citep{harutyunyan2017multitask}, we consider adult admissions with a single, unique ICU visit. This excludes patients with transfers between different ICUs. Patients without labels or observations in the ICU were excluded, as were patients who died or were discharged before 48 hours. After applying exclusion criteria, our final dataset included 21,139 patient admissions and 2,797 deaths.

We used the same feature extraction procedure as detailed in \citep{harutyunyan2017multitask}. Code to generate these data are publicly available\footnote{https://github.com/YerevaNN/mimic3-benchmarks}. For completeness, we briefly describe the feature extraction process here. For each admission, we extracted 17 features (\textit{e.g.}, heart rate, respiratory rate, Glasgow coma scale) from the first 48 hours of their ICU visit. We applied mean normalization and discretization, resulting in 59 features. Sampling rates were set uniformly to once per hour using carry-forward imputation. Mask features, indicating if a value had been imputed resulted in an additional 17 features. After preprocessing, each example was represented by $d=76$ time-series of length $T=48$ and a binary label indicating whether or not the patient died during the remainder of the hospital visit.

Given these data, the goal is to learn a mapping from the features to the probability of in-hospital mortality, resulting in a single prediction per patient admission. We measured performance by calculating the area under the receiver operating characteristic curve (AUROC) and the area under the precision recall curve (AUPR). We randomly split the data into training (70\%), validation (15\%), and testing (15\%): 14,681 (1,987 deaths) in training, 3,222 (436 deaths) in validation and 3,236 (374 deaths) in test. We learned model parameters and selected hyperparameters using training and validation data and evaluated model performance using held-out test data. Specifics on hyperparameter search are presented in Section  \ref{implementation_details}. We generated empirical 95\% confidence intervals by bootstrapping the test set. 

\subsection{Baseline CNN Architecture}\label{sec:basecnn}
\begin{figure}
    \centering
    \includegraphics[width=.75\linewidth]{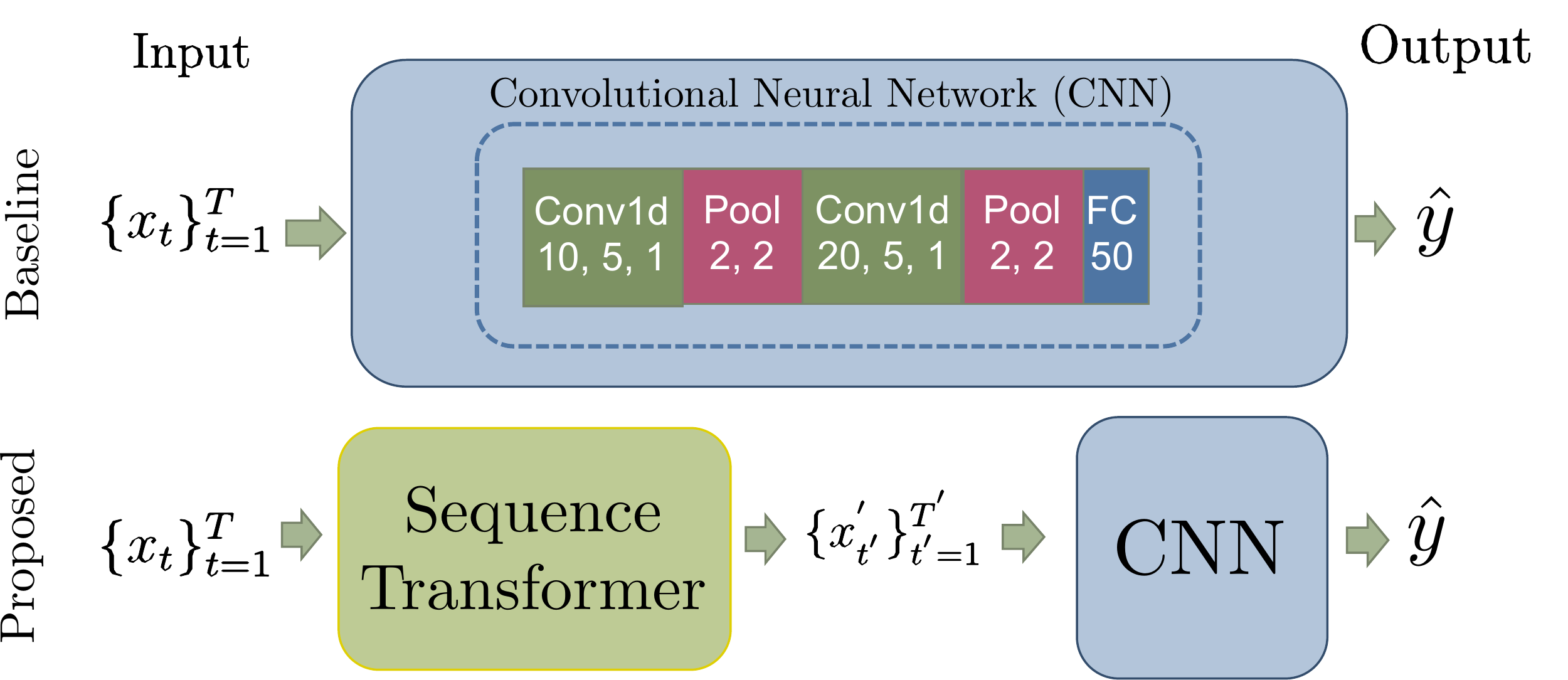}
    \caption{The architecture of the CNN. Baseline CNN inputs ($\{x_t\}_{t=1}^{T}$) or Sequence Transformer inputs ($\{x^{'}\}_{t^{'}=1}^{T^{'}}$) are fed into a standard CNN that outputs our in-hospital mortality prediction. Here, the admission indexing $i$ is omitted for simplicity. Convolutional and maxpool layers are annotated with the number of outputted channels (omitted for maxpool), 1D filter size and stride. Fully connected layers (FC) are annotated with the number of neurons.}
    \label{fig2}
\end{figure}

As a baseline with which to compare, we considered a CNN without any additional Sequence Transformer. We compared the discriminative performance of a CNN with original inputs to a CNN with inputs transformed via the Sequence Transformer. We referred to the first method as our Baseline CNN. The second is our proposed method: Sequence Transformer Networks. The only difference between this baseline and our proposed approach is the Sequence Transformer (Figure \ref{fig2}). Both models feed either the original or transformed example into a standard 1D CNN. For this CNN, we used the two layer CNN described in Figure \ref{fig2}. The CNN consists of two 1D convolutional and pooling layers followed by a single, hidden, fully connected layer.

In addition to considering a baseline consisting of no transformations, we also considered networks that used either i) temporal transformations only or ii) magnitude transformations only. This allowed us to measure the marginal contribution of each transformation in the Sequence Transformer.

\subsection{Implementation Details} \label{implementation_details}

We optimized the following hyperparameters: network depth, number of neurons in the final fully connected hidden layer, batch size, and dropout rate. We trained twenty models with randomly selected hyperparameters, for at most 10 epochs. Hyperparameters were randomly chosen from predefined sets of values. Batch size was randomly selected from: 8, 15, 30. The rate of dropout was randomly selected from: 0, .1, .2, …, .9. We tested CNN architectures of depth 2, 3 and 4. Finally, the number of neurons in the final fully connected hidden layer was randomly chosen from: 50, 100, 250 and 500. The settings that led to the best performance on the validation data are shown in Figure \ref{fig2}.

Since these hyperparameters were tuned for our Baseline CNN using the original input, we also considered a model tuned to the transformed signal. The resulting optimal hyperparameters were largely unchanged, except that we found that a dropout rate of 0.2 (vs. 0.3) worked better for Sequence Transformer Networks. The optimal batch size for both models was 15. 

During model training, we included gradient clipping. This consisted of a reduced slope from 1 to .01 outside of a reasonable range of transformation parameter values. In practice, we set this range to $[-2,2]$. We found this implementation detail to be important. Without it, we witnessed quick increases in the value of the transformation parameters that led to unrecoverable model states. 

\section{Results}

We present the performance of the Baseline CNN, which takes as input untransformed signals as described in Section \ref{sec:basecnn}, vs. Sequence Transformer Networks. We further break down the Sequence Transformer into its two parts: temporal and magnitude transformations and evaluate their individual contributions. Finally, we investigate the learned transformations through a series of visualizations and analyze the effect of Sequence Transformer Networks on intra-class signal similarity.

\subsection{CNN Baseline vs Sequence Transformer Networks}
Our proposed method, Sequence Transformer Networks, outperforms the Baseline CNN, in terms of both AUROC and AUPR, on the task of predicting in-hospital mortality using data from the first 48 hours (Table \ref{tbl1}). 

\begin{table}[]
\centering
\caption{Test performance for the task of predicting in-hospital mortality. Relative to the baseline performance, transforming the input before feeding it into the CNN results in consistent improvements in both the area under the receiver operating characteristics curve (AUROC) and the area under the precision recall curve (AUPR).}
\label{tbl1}
\begin{tabular}{@{}lll@{}}
\toprule
Method & AUROC (95\% CI)& AUPR (95\% CI)\\ \midrule
Baseline CNN & 0.838 (0.820, 0.859) & 0.445 (0.393, 0.495)\\
Sequence Transformer Networks&\textbf{0.851} (0.833, 0.871)&\textbf{0.476} (0.424, 0.527)\\
\hspace{5mm}Temporal Transformations Only & 0.846 (0.827, 0.867)& 0.452 (0.393, 0.500)\\
\hspace{5mm}Magnitude Transformations Only & 0.846 (0.826, 0.867)& 0.463 (0.408, 0.516)\\ \bottomrule
\end{tabular}
\end{table}

\begin{figure}
\centering
    \begin{subfigure}[b]{0.5\textwidth}            
            \includegraphics[width=\textwidth]{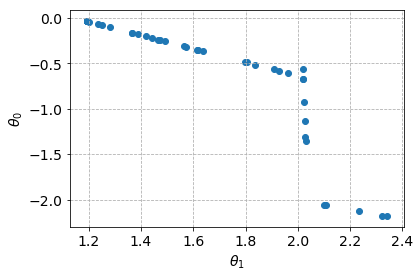}
            \caption{}
            \label{fig:7i_lr}
    \end{subfigure}%
    \begin{subfigure}[b]{0.5\textwidth}
            \centering
            \includegraphics[width=\textwidth]{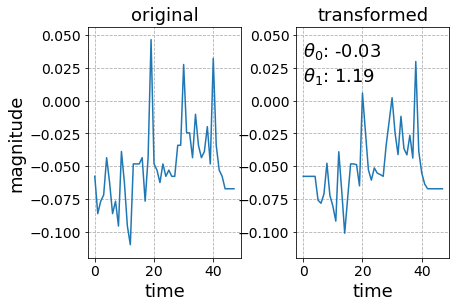}
            \caption{}
            \label{fig:7i_transform}
    \end{subfigure}
    \caption{Sequence Transformer Networks: Temporal Transformations Only. \textbf{(a)} Visualization of temporal transformation parameters applied to the test set. Note that $\theta_{1} \geq 0$ indicates signal compression, while $\theta_{0} \leq 0$ indicates shifting the signal forward in time.  \textbf{(b)} A random test patient's normalized diastolic blood pressure before and after $\boldsymbol{\theta}$ transformation ($\theta_1=1.19$, $\theta_0=-0.03$). In addition to signal compression and shifting, the network smooths the signal.}\label{fig:7ijoh}
\end{figure}

Compared to the Baseline CNN, Sequence Transformer Networks incorporates a secondary, transformation network. However, the improvement in performance is not due to the additional complexity of the model. For both models, we tuned the depth of the CNN architecture. In both cases, the best CNN, determined by validation performance and presented in the results, had a network depth of 2. Therefore a deeper network alone is not sufficient for increasing performance. 

Since the Sequence Transformer consists of two transformations, we further break down the performance increase into: temporal transformations and magnitude transformations. In Table \ref{tbl1}, we see that both types of transformations lead to marginal improvements over the baseline. Moreover, their combination appears to be complementary, though the difference is small. 

\begin{figure}
\centering
    \begin{subfigure}[b]{0.5\textwidth}            
            \includegraphics[width=\textwidth]{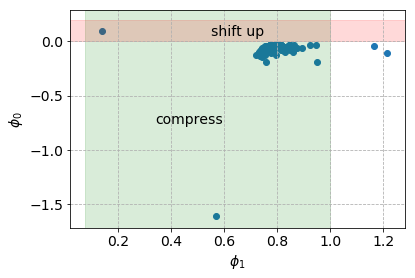}
            \caption{}
            \label{fig:SRl}
    \end{subfigure}%
    \begin{subfigure}[b]{0.5\textwidth}
            \centering
            \includegraphics[width=\textwidth]{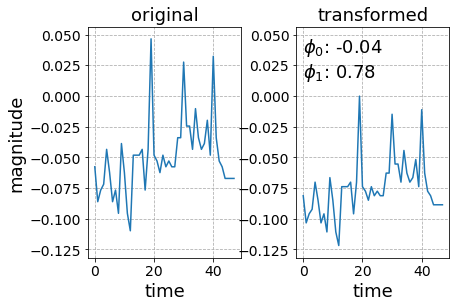}
            \caption{}
            \label{fig:D-Imager}
    \end{subfigure}
    \caption{Sequence Transformer Networks: Magnitude Transformations Only. \textbf{(a)} Visualization of magnitude transformation parameters applied to the test set. Note that $\phi_{1} \leq 1$ indicates signal value compression, while $\phi_{0} \leq 0$ indicates a downward shift.  \textbf{(b)} A random test patient's normalized diastolic blood pressure before and after $\boldsymbol{\phi}$ transformation ($\phi_1=0.78$, $\phi_0=-0.04$).}\label{fig:12}
\end{figure}

\subsection{Learned Temporal and Magnitude Transformations}

In this section, we qualitatively explore what the Sequence Transformer has learned. Figure \ref{fig:7ijoh} summarizes the transformation learned using a network that employs only temporal transformations. Recall that the transformation depends on the input. Figure \ref{fig:7i_lr} shows the empirical distribution of the two temporal transformation parameters ($\theta_1$, $\theta_0$). Each point represents a temporal transformation learned for a specific patient admission in the test set. In this case, most of the data occur around $\theta_1=1.19$ and $\theta_0=-0.03$. Essentially, the network learns to compress the original signal ($\theta_{1} \geq 1$) and shift the signal forward in time ($\theta_{0} \leq 0$) by various degrees. In doing so, the network learns how to align the time-series data from different patient admissions. Figure \ref{fig:7i_transform} shows the original and the temporally transformed normalized diastolic blood pressure for a randomly selected patient in the test set. In line with the results shown in the previous figure, the signal is compressed along the x-axis and shifted forward in time. In Figure \ref{fig:7i_transform}, though the signal is moved forward in time, it is not clipped, but rather compressed. This suggests that $\theta_0$ is helping to center the signals. The sudden drop off at $\theta_1=2$ is most likely due to the gradient clipping, since that is where it begins to take effect. In addition, we observe a smoothing effect that is due, in part, to the the linear interpolation.

Figure \ref{fig:12}, shows the same type of plots as Figure \ref{fig:7ijoh} but for a network that includes only magnitude transformations. We observe that the signal is, on average, shifted down and compressed. Similar to the temporal transformations, the magnitude transformations help align signals. Amplitude and offset invariances have a clinical significance for many features in this dataset including blood pressure, heart rate, respiratory rate and temperature. We hypothesize that these transformations help account for different physiological baselines.

\begin{figure}
\centering
    \begin{subfigure}[b]{0.32\textwidth}            
            \includegraphics[width=\textwidth]{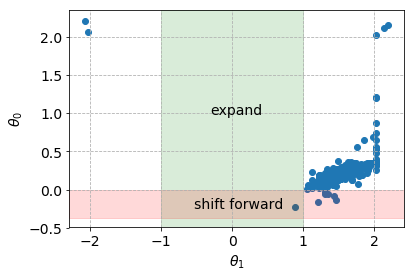}
            \caption{}
            \label{fig:8j_ud1}
    \end{subfigure}%
    \begin{subfigure}[b]{0.32\textwidth}
            \centering
            \includegraphics[width=\textwidth]{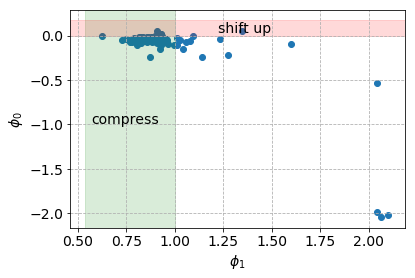}
            \caption{}
            \label{fig:8j_ud2}
    \end{subfigure}
    \begin{subfigure}[b]{0.32\textwidth}
            \centering
            \includegraphics[width=\textwidth]{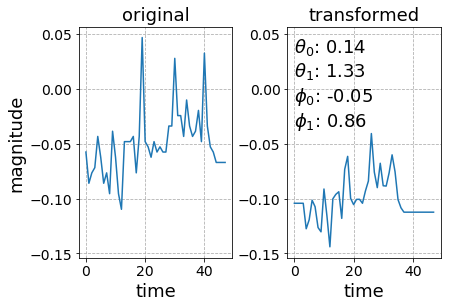}
            \caption{}
            \label{fig:8j_transform}
    \end{subfigure}
    \caption{Sequence Transformer Networks  \textbf{(a)} Visualizations of temporal transformation parameters applied to the test set. On average, network compresses and shifts signals backwards in time. \textbf{(b)} Visualizations of magnitude transformation parameters applied to the test set. On average signal values are compressed and shifted down. \textbf{(c)} The learned network smooths and shifts the normalized diastolic blood pressure to the left bottom direction of the frame for a randomly selected patient using the transformations: $\theta_1=1.33$, $\theta_0=0.14$, $\phi_1=0.86$ and $\phi_0=-0.05$.}\label{fig:8j}
\end{figure}

Finally, we visualize the output of the Sequence Transformer, which learns temporal, amplitude and offset invariances together (Figure \ref{fig:8j}). In Figures \ref{fig:8j_ud1} and \ref{fig:8j_ud2}, each point represents a transformation learned for a specific patient in the test set. We see that the network, on average, compresses the signal and shifts it slightly back in time. In the temporal transformation only network (Figure \ref{fig:7ijoh}), the network shifted signals forward in time. This suggests that the direction of the shift is less important than the overall alignment of the different patients. For magnitude transformations, the network on average compresses the signal and shifts it down. These learned transformation trends align with the magnitude transformation trends learned separately (Figure \ref{fig:12}). In Figure \ref{fig:8j_transform} we illustrate the transformations applied to a random test patient's normalized diastolic blood pressure. 

\subsection{Increasing Intra-Class Similarity}

Sequence Transformer Networks have the ability to learn transformations that reduce label independent variations in the signal. By reducing irrelevant variance, transformed signals from patients with similar outcomes then appear more similar. We investigate this property by analyzing the intra-class Euclidean pairwise distance. On each dataset (original vs. transformed), we calculated the Euclidean pairwise distance between admissions labeled positive and the Euclidean pairwise distance between those labeled negative. 

The transformed dataset had on average lower pairwise intra-class distances compared to the original (untransformed) data  (positive: 28.2 vs. 34.9 and negative: 26.3 vs. 31.8). We hypothesize that this increase in intra-class similarity contributes to the overall improved discriminative performance of the  Sequential Transformer Network over the Baseline CNN. 

\section{Conclusion}
In this paper, we proposed the use of an end-to-end trainable method for exploiting invariances in clinical time-series data. Building off of ideas first presented in the context of transforming images, we extended the capabilities of CNNs to capture temporal, amplitude, and shift invariances. In general, such invariances may be task dependent (\textit{i.e.}, may depend on the outcome of interest or the population studied). Given the large number of possible clinical tasks, techniques that automatically learn to exploit invariances based on the data have a clear advantage over preprocessing techniques. 

We demonstrated that this method leads to improved discriminative performance over the Baseline CNN, on the task of predicting in-hospital-morality from multivariate clinical time-series data collected during the first 48 hours of an ICU admission. Though the difference in performance is small, the improvement is evident across both AUROC and AUPR. 

The proposed approach is not without limitation. More specifically, in its current form the Sequence Transformer applies the same transformation across all features within an example, instead of learning feature-specific transformations. Despite this limitation, the learned transformations still lead to an increase in intra-class similarity. In conclusion, we are encouraged by these preliminary results. Overall, this work represents a starting point on which others can build. In particular, we hypothesize that the ability to capture local invariances and feature-specific invariances could lead to further improvements in performance.

\acks{This work was supported by the National Science Foundation (NSF award no. IIS-1553146) and the National Institute of Allergy and Infectious Diseases of the National Institutes of Health (grant no. U01AI124255). The views and conclusions in this document are those of the authors and should not be interpreted as necessarily representing the official policies, either expressed or implied, of the National Science Foundation nor the National Institute of Allergy and Infectious Diseases of the National Institutes of Health.}

\bibliography{main}
\end{document}